# SIMPLE TEXT MINING FOR SENTIMENT ANALYSIS OF POLITICAL FIGURE USING NAÏVE BAYES CLASSIFIER METHOD

Yustinus Eko Soelistio *, Martinus Raditia Sigit Surendra †
System Information, Faculty of Information and Communication Technology, Multimedia Nusantara University
Jl.Scientia Boulevard, Gading Serpong, Tangerang, Banten-15811, Indonesia
email : * yustinus.eko@umn.ac.id, † sigit@umn.ac.id

## ABSTRACT

Text mining can be applied to many fields. One of the application is using text mining in digital newspaper to do politic sentiment analysis. In this paper sentiment analysis is applied to get information from digital news articles about its positive or negative sentiment regarding particular politician.

This paper suggests a simple model to analyze digital newspaper sentiment polarity using naïve Bayes classifier method. The model uses a set of initial data to begin with which will be updated when new information appears. The model showed promising result when tested and can be implemented to some other sentiment analysis problems.

## 1. INTRODUCTION

Indonesia is one of the big democratic nation. Almost everyday there are news about politician that cover many topics include corruption and regional election. Mass media has important role in delivering news therefore can influence public opinion. For example, one news media give positive review on one candidate while others give negative one.

Nowadays news media can deliver their content through digital media. This accessibility opens new opportunity to analyze news content with text mining. Digital news media can be considered as unstructured data. This huge amount of data available on the web creates today an information overloading problem [5].

Text mining has been implemented in many applications such as [1,2,3,4,6,8]. One of the suggested implementation is for analyzing readers' sentiment on some particular news. Research result by [1] suggest that naïve Bayesian classifier and support vector machines can be used to identify readers opinion, either positive or negative, on English movie's review and Indonesian daily news.

This paper suggests a method to implement sentiment's analysis using naïve Bayesian method on digital articles and newspapers. The sentiment analysis focuses on the probability of whether news media give positive or negative review on some particular political figures.

## 2. BASIC MODEL AND ASSUMPTION

The model starts from what has been suggested in [2] that consider "who" is speaking, "to whom" is speaking, and "what" as variables. This paper adopts those variables and uses them to determined sentiment probability. The values of those variables are updated according to what the system has learned from training data $T$. Variables "who" ($w$), "whom" ($h$) and "what" ($a$) store their values in a form of matrix $M$ and $N$ as knowledge base set. $M$ is used to stores prior polarity ($p$) of $w$ towards $h$, and $N$ stores how many times $w$ give such statement toward $h$ ($s_{(w,h)}$). The default values are $\forall(p_{(w,h)}, s_{(w,h)}) = 0$. These values change with $T$ so when

$$\begin{cases} (p_{(w,h)} \wedge s_{(w,h)} > 0) \rightarrow likely\ positive\ sentiment \\ (p_{(w,h)} \vee s_{(w,h)} = 0) \rightarrow likely\ neutral\ sentiment \\ (p_{(w,h)} \wedge s_{(w,h)} < 0) \rightarrow likely\ negative\ sentiment \end{cases}$$
(1)

Training data $T$ is a list of independent articles $C_i$. Each articles can contains one or more political figures "keyword", therefore

$$|(w, h, a) \in C_i| \geq 0 \qquad (2)$$

The sentiment of article $C_i$ is cast by $p_{(w,h)}$ and $s_{(w,h)}$ in $C_i$. Each $p_{(w,h)}$ is determined by the value of $a$ which correspond to unique word $o$ in database $D$. $w$, $h$ and $o$ are handled as a token like in [3,7]. $o$ can be "negative" words ($-o$) such as corruption, convict, and dispute, or "positive" words ($+o$) like honest, improve, and hope. Each appearance of $p_{(w,h,a)}$ will also increase value of $s_{(w,h)}$ by one. If each $o \in D$ have a value of integer $a$ then

$$\begin{cases} -o \rightarrow (a = -1) \rightarrow (p_{(w,h,-o)} < 0) \land (s_{(w,h)} > 0) \\ +o \rightarrow (a = +1) \rightarrow (p_{(w,h,+o)} > 0) \land (s_{(w,h)} > 0) \end{cases}$$
(3)

Values of $p_{(w,h,a)}$ are stored in $w \times h$ matrices of matrix $M$ and value of $s_{(w,h)}$ are stored in $w \times h$ matrices of matrix $N$. $w$ is the "who" where the statement come from in a article $C_i$, and $h$ is the "to whom" or "to who" $w$ give his/her statement to. Since there are many combinations of structures in a sentence then seven assumptions will be set and used through out this paper.

### Assumption 1

There are only two types of articles in the news, first is articles which discuss about one or more politician, and second articles that do not say anything about politician (even though the article is about politic). This assumption will hold true equation (2) since all articles that discuss one or more politician will have $|(w,h,a) \in C_i| > 0$ and the others $|(w,h,a) \in C_i| = 0$.

### Assumption 2

For each statement $o$ there are always person $w$ who declares, and person $h$ whose the $o$ are declared to. Thus whether exist $o$ and $w$ then there is always $h$, and whether exist $o$ and $h$ then there is always $w$. This assumption makes sure that change in $p_{(w,h)}$ will always by the value of $a$, and $a$ always have references to $p_{(w,h)}$.

### Assumption 3

Though assumption 2 will hold for most statements $o$ in $C_i$, there are some possibilities that it will not. There are some cases where there is no reference of $w$ but $o$ is present, such may happen in the first sentence of $C_i$. Assumption 3 will guarantee that assumption 2 will always be true by assigning $w$ to the news media where the article appears, hence the default value of $w = news\ media$.

### Assumption 4

Article $C_i$ can have two or more $w$ and $h$ therefore the system keep track of $w_y$ and $h_y$ by changing their values to the most recent politician keyword found. For example let say $b_z$ is words in $C_i$ then $b_z \in C_i$ and

$$\begin{cases} (b_z = who\ keyword) \land (b_z \neq w_y) \rightarrow w_{y+1} = b_z \\ (b_z = whom\ keyword) \land (b_z \neq h_y) \rightarrow h_{y+1} = b_z \end{cases}$$
(4)

This assumption ensure that each $o$ give the correct $a$ to $p_{(w,h)}$.

### Assumption 5

Every negation keyword ($g$) in a sentence such as "no" and "not" will change the polarity of $o$ thus change value of $a = -1$ for $p_{(w,h,-o)}$ to $a = +1$. If there exist two or more $g$ in one sentence then polarity of $o$ will be changed as many times as $g$ appear.

### Assumption 6

To distinguish between $+o$ and sarcasm, the system check $+o$ with the value of prior $p_{(w,h)}$. If $p_{(w,h)} < 0$ then the $+o$ will be considered as sarcasm, otherwise it will be considered as legitimate positive statement. Polarity of $+o$ will not be changed thus $a = +1$ will be added to $p_{(w,h)}$.

### Assumption 7

This paper assumes that all news media have some sentiment tendency towards politician. Therefore each $o$ in $C_i$ which correspond to $h$ change the probable polarity of news media towards $h$. Thus for every $p_{(w,h,a)}$ will change $p_{(w,h)}$ with the same value of $a$ and increase $s_{(w,h)}$ by one.

## 3. TRAINING SET AND NAIVE BAYESIAN MODEL

Training set $T$ consists of $C_i$ therefore $T \ni w, h, a$ and $a = +1 \leftarrow -o$ or $a = +1 \leftarrow +o$. To find out whether a news media give positive or negative review on particular politician, this paper adapts Bayesian model from [1,4,11] and modify it so

$$P(p_{(w,h)}|s_{(w,h)}) = \frac{p_{(w,h)}}{s_{(w,h)}} \quad (5)$$

where $P(p_{(w,h)}|s_{(w,h)})$ is the probability of $p_{(w,h)}$ in given event $s_{(w,h)}$. Since $p_{(w,h)}$ can be either has positive value or negative value and $|p_{(w,h)}| \leq s_{(w,h)}$ then $-1 \leq P(p_{(w,h)}|s_{(w,h)}) \leq 1$. When $P(p_{(w,h)}|s_{(w,h)}) \sim -1$ then $w$ tends to give negative review on $h$, where $P(p_{(w,h)}|s_{(w,h)}) \sim 1$ states otherwise. For example when $P(p_{(w,h)}|s_{(w,h)}) = -0.95$ means that $w$ has 95% probability to has negative sentiment towards $h$. Thus the probability of sentiment polarity of article $C_i$ towards $h$ is

$$P_{c_{(i,h)}} = \sum_{y=1}^{y=n} p_{(w_{(i,y)},h_i)} / \sum_{y=1}^{y=n} s_{(w_{(i,y)},h_i)} \quad (6)$$

Following equation (3) then positive value $a$ will add $p_{(w,h)}$ value, and negative $a$ will subtract $p_{(w,h)}$ value. Hence equation (1) concludes that the further away value $p_{(w,h)}$ from 0 then the higher probability of $w$ has $o$ sentiment (either positive or negative) towards $h$.

|       | $w_{(1)}$ | $w_{(2)}$ | $w_{(3)}$ | $w_{(4)}$ | $w_{(5)}$ | $w_{(q)}$ |
|-------|-------|-------|-------|-------|-------|-------|
| $h_{(1)}$ | p(1,1) | p(2,1) | p(3,1) | p(4,1) | p(5,1) | p(n,1) |
| $h_{(2)}$ | p(1,1) | p(2,2) | p(3,2) | p(4,2) | p(5,2) | p(n,2) |
| $h_{(3)}$ | p(1,1) | p(2,3) | p(3,3) | p(4,3) | p(5,3) | p(n,3) |
| $h_{(4)}$ | p(1,1) | p(2,4) | p(3,4) | p(4,4) | p(5,4) | p(n,4) |
| $h_{(5)}$ | p(1,1) | p(2,5) | p(3,5) | p(4,5) | p(5,5) | p(n,5) |
| $h_{(r)}$ | p(1,n) | p(2,n) | p(3,n) | p(4,n) | p(5,1) | p(n,n) |

Figure 1. Structure of matrix $M$

Initial values of all cells in figure 1 are 0. The values will change after the system goes through $T$. Each cell stores historical data of sentiment $w$ towards $h$. Number of $w$ and $h$ do not have to be equal. It is possible that $w_q = h_r$ in a case of $w$ give a statement about him/her self.

Matrix $N$ has the same structure as matrix $M$ with the difference $p_{(w,h)}$ cells. In matrix $N$, values of $p_{(w,h)}$ are substituted with values of $s_{(w,h)}$.

## 4. SYSTEM DESIGN AND TEST MODEL

This paper creates a system design to test the model. The test will use articles $C$ from Indonesia's digital newspaper $K$ to create $T$. To test the accuracy, the system will try to determine $P(p_{(w,h)}|s_{(w,h)})$ from another article $C \notin T$. Set of $o$ and $a$ are determined beforehand by human interpreter.

Before the test, a general system flow is established. The general system flow consists of five modules which are reader and parser, cleanser, helper, analyzer, and display.

Reader and parser separate each words and punctuation mark from the text. Then these words will be filtered by cleanser. The filter removes special words and adverbs from the sentence by comparing them with a set of pre define filter words such as '*seorang*' ('a', 'an' in English) ,'*adalah*' ('is' in English), '*yang*' ('that' in English). As example a sentence like '*ABC adalah seorang koruptor*' (ABC is an corruptor) will become '*ABC koruptor*'.

Sentences that have been cleansed will be scanned by helper to find non-common pronoun or politician names such as alias or pronoun. This module will change those words and replace them with system's keywords. These keywords will be used by analyzer to locate $w$, $h$, and $o$.

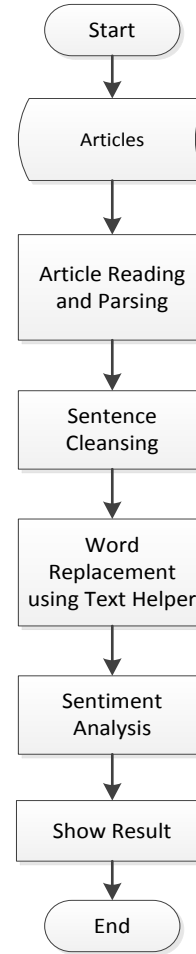

Figure 2. System's activity diagram

The system reads all sentences separately hence all sentences are evaluated independently. Analyzer process will be triggered by a condition where the system found special word. These special words can be categorized into negative and positive types.

Figure 3 is process flow when the system found negative word like *koruptor* ('corruptor' in English), and *tersangka* ('suspect' in English). The result of process in figure 3 are $w$, $h$ and $p_{(w,h)}$. Using this result system can create matrix $M$ and $N$.

The process flow of positive type is similar to the process flow of negative type. The only different is that the keywords $o$ and the opposite value (+1) to be assigned for positive words.

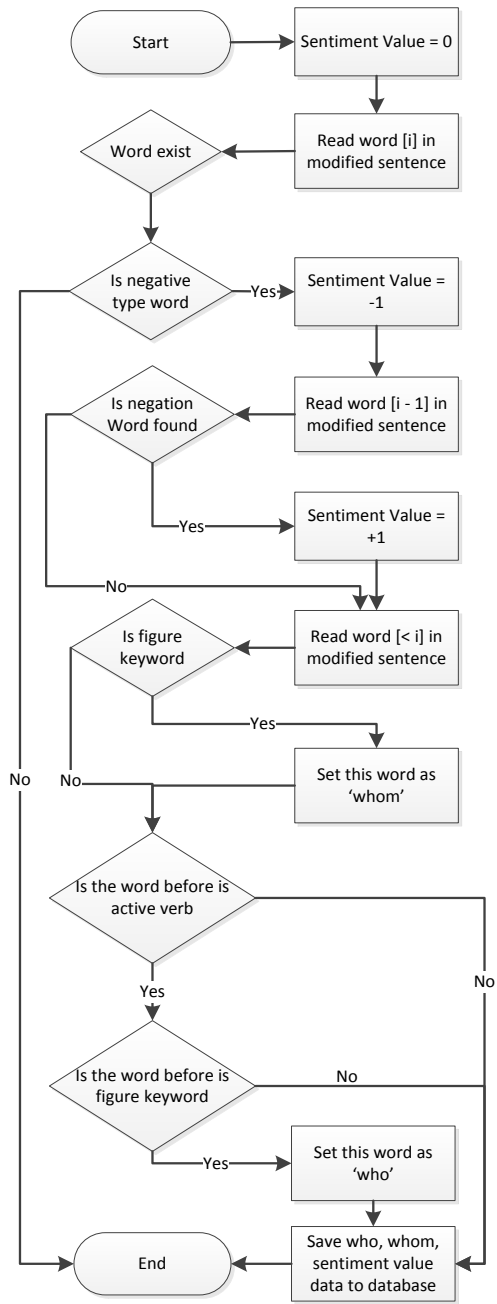

Figure 3. The process flow of negative type words scenario.

Figure 4 shows the result of $P(p_{(kpk,andi)}|s_{(kpk,andi)}) = \frac{-1}{1}$, $P(p_{(K,andi)}|s_{(K,andi)}) = \frac{-6}{6}$, $P(p_{(K,kpk)}|s_{(K,kpk)}) = \frac{1}{1}$, and $P(p_{(kpk,deddy)}|s_{(kpk,deddy)}) = \frac{-1}{1}$. By employing formula (6) then $P_{C_{(i,andi)}} = -1$, $P_{C_{(i,kpk)}} = 1$, and $P_{C_{(i,deddy)}} = -1$. Thus it means that this particular article from $K$ has probability of 100% giving negative sentiment to Andi and Deddy, and has probability of 100% giving positive sentiment to $kpk$. These result will be saved as system's knowledge base $T$ and will be used to analyze $C_{i+1}$.

| Article Index | Sentence Index | Who | Whom | Value |
|---|---|---|---|---|
| 1 | 1 | kpk | andi | -1 |
| 1 | 5 | 0 | andi | -1 |
| 1 | 7 | 0 | andi | -1 |
| 1 | 8 | 0 | andi | -1 |
| 1 | 8 | 0 | kpk | 1 |
| 1 | 10 | 0 | andi | -1 |
| 1 | 12 | 0 | andi | -1 |
| 1 | 14 | 0 | andi | -1 |
| 1 | 17 | kpk | deddy | -1 |

Figure 4. Result after running data $T$

Table 1. Matrix $M$ after running data $T$

|  | $w_{(K)}$ | $w_{(kpk)}$ |
|---|---|---|
| $h_{(K)}$ | 0 | 0 |
| $h_{(kpk)}$ | 1 | 0 |
| $h_{(andi)}$ | -7 | -1 |
| $h_{(deddy)}$ | 0 | -1 |

Table 2. Matrix $N$ after running data $T$

|  | $w_{(K)}$ | $w_{(kpk)}$ |
|---|---|---|
| $h_{(K)}$ | 0 | 0 |
| $h_{(kpk)}$ | 1 | 0 |
| $h_{(andi)}$ | 7 | 1 |
| $h_{(deddy)}$ | 0 | 1 |

The algorithm is tested using Java and MySQL database to save the data. The test uses articles from $K$ digital newspaper as training data $T$ and evaluated articles $C_i$.

From $T$, the system form matrix $M$ and $N$ with $w = kpk, 0$ and $h = andi, kpk, deddy$. $w = 0$ is the default value which equals to $w = K$.

Figure 5 shows the result after running $C_{(i+1)}$, $P(p_{(K,andi)}|s_{(K,andi)}) = \frac{-1}{1}$, $P(p_{(km,andi)}|s_{(km,andi)}) = \frac{1}{1}$, and $P(p_{(ahmad,andi)}|s_{(ahmad,andi)}) = \frac{2}{2}$,

Following formula (6) then $P_{C_{(i+1,andi)}} = \frac{2}{4}$. Thus it means that article $C_{(i+1)}$ has probability of 50% giving positive sentiment to Andi.

Figure 5. Result after running $C_{(i+1)}$

Table 3. Matrix $M$ after running $C_{(i+1)}$

|  | $w_{(K)}$ | $w_{(km)}$ | $w_{(ahmad)}$ |
|---|---|---|---|
| $h_{(andi)}$ | -1 | 1 | 2 |

Table 4. Matrix $N$ after running $C_{(i+1)}$

|  | $w_{(K)}$ | $w_{(km)}$ | $w_{(ahmad)}$ |
|---|---|---|---|
| $h_{(andi)}$ | 1 | 1 | 2 |

$P_{C_{(i+1,andi)}}$ will be calculated proportional to $\langle P_{C_{(i,andi)}} \rangle$ subsequently $P_{T_{(andi)}} = \frac{\sum_1^n P_{C_{(i,andi)}}}{n} = \frac{-1+\frac{1}{2}}{2}$. Then it signifies $K$ has 25% probability to have negative sentiment to Andi. Thus in general we get

$$P_{T_{(h)}} = \frac{\sum_1^n P_{C_{(i,h)}}}{n} \qquad (7)$$

## 5. CONCLUSION

The model proposed in this paper can be used to analyze sentiment of an article in digital news media towards politician. The algorithm in the model is fairly general to be used in many other sentiment analysis problems with only few modifications. The prompt test shows promising results, nevertheless further test with bigger data set and more complex slang language yet has to be done.

One major problem in applying this model is to accurately identify the 'who' and the 'whom' in the articles on precise context. Finding meaning in context behind each sentence and associate it with whole articles' context is still a challenge for the future.